\documentclass{article}

\usepackage{arxiv}

\usepackage[utf8]{inputenc} 
\usepackage[T1]{fontenc}    
\usepackage{hyperref}       
\usepackage{url}            
\usepackage{booktabs}       
\usepackage{amsfonts}       
\usepackage{nicefrac}       
\usepackage{microtype}      
\usepackage{lipsum}
\usepackage{graphicx}
\usepackage{authblk}
\usepackage{amsmath}

\usepackage{times}
\usepackage{latexsym}
\usepackage{xcolor}
\usepackage{pifont}
\usepackage[misc]{ifsym}
\usepackage{hyperref}
\usepackage{url}
\usepackage{color,array}
\usepackage{cite}
\usepackage{graphicx}
\usepackage{tabularx}
\usepackage{multirow}
\usepackage{lmodern}
\usepackage[most]{tcolorbox}
\usepackage{fontawesome}
\usepackage{fancyvrb}
\usepackage{comment}
\usepackage{float}
\usepackage{longtable}
\usepackage{amsmath}
\usepackage{amssymb}
\usepackage{booktabs}

\title{Fake News Detection: It's All in the Data!}

\author{
    Soveatin Kuntur\textsuperscript{*} \textit{Corresponding Author},
    Anna Wróblewska\textsuperscript{**},
    Marcin Paprzycki\textsuperscript{***},
    Maria Ganzha\textsuperscript{****}
    \\
    \textit{Warsaw University of Technology, plac Politechniki 1, 00-661 Warsaw, Poland}\\
    \textsuperscript{*}\href{mailto:soveatin.kuntur.dokt@pw.edu.pl}{soveatin.kuntur.dokt@pw.edu.pl},
    \textsuperscript{**}\href{mailto:anna.wroblewska1@pw.edu.pl}{anna.wroblewska1@pw.edu.pl},
    \textsuperscript{****}\href{mailto:maria.ganzha@pw.edu.pl}{maria.ganzha@pw.edu.pl}
    \\
    \textit{Systems Research Institute, Polish Academy of Sciences, Newelska 6, 01-447 Warsaw, Poland}\\
    \textsuperscript{***}\href{mailto:marcin.paprzycki@ibspan.waw.pl}{marcin.paprzycki@ibspan.waw.pl}
}

\begin{document}
\maketitle
\begin{abstract}
This comprehensive survey serves as an indispensable resource for researchers embarking on the journey of fake news detection. By highlighting the pivotal role of dataset quality and diversity, it underscores the significance of these elements in the effectiveness and robustness of detection models. The survey meticulously outlines the key features of datasets, various labeling systems employed, and prevalent biases that can impact model performance. Additionally, it addresses critical ethical issues and best practices, offering a thorough overview of the current state of available datasets. Our contribution to this field is further enriched by the provision of a \href{https://github.com/fakenewsresearch/dataset}{GitHub repository}, which consolidates publicly accessible datasets into a single, user-friendly portal. This repository is designed to facilitate and stimulate further research and development efforts aimed at combating the pervasive issue of fake news.
\end{abstract}


\section{Introduction}
Detecting fake news is essential in today's digital era due to its profound impact on individuals, societies, and democratic processes \cite{lazer2018science, lewandowsky2017beyond, reisach2021responsibility}. Identifying and debunking false information ensures the dissemination of accurate and reliable information, protecting public discourse from manipulation and mitigating the harm caused by rumors, conspiracy theories, and false narratives~\cite{shu2019beyond, abdelminaam2021coaid, goldani2021detecting, reisach2021responsibility, truicua2023s}. Effective fake news detection maintains trust in media sources, promotes critical thinking, and prevents manipulation by malicious actors. It is also crucial for cybersecurity, as misinformation can facilitate the spread of malware and phishing attacks. Addressing fake news thus supports social cohesion and upholds ethical standards in digital communication, safeguarding democratic processes from manipulation and propaganda.

Given the critical role of fake news detection in preserving information integrity, it is important to examine the development of these detection systems, particularly the use of datasets. High-quality and diverse datasets are crucial for capturing various misinformation patterns, enhancing the effectiveness of detection models that integrate textual, visual, and behavioral features~\cite{peng2024not, jain2024confake, lai2024rumorllm}.

This survey investigates the key components of datasets used in developing fake news detection models. It explores the characteristics, common features, and labels of existing datasets, evaluating their impact on the effectiveness and resilience of detection algorithms. Emphasis is placed on the importance of collecting representative and reliable datasets. The survey also addresses the challenges, biases, and ethical considerations associated with these datasets, the role of multimodal datasets, and best practices for constructing high-quality datasets. Additionally, it reviews the evolution of fake news detection models with the availability of diverse datasets and considers future directions for advancing detection technologies.
\section{Characteristics of existing fake news datasets}

In this section, we contribute by collecting publicly available datasets, summarizing their contents, and comparing them on our GitHub page\footnote{\href{https://github.com/fakenewsresearch/dataset}{https://github.com/fakenewsresearch/dataset} - This link is currently private}. This initiative offers researchers a centralized, comprehensive portal for accessing and analyzing relevant datasets, with regular updates. Due to page constraints, only a portion of the GitHub pages are displayed here. Table~\ref{tab:datasets} provides an overview of key datasets included in the repository.

\section{Characteristics of existing fake news datasets}

In this section, we contribute by collecting publicly available datasets, summarizing their contents, and comparing them on our GitHub page\footnote{\href{https://github.com/fakenewsresearch/dataset}{https://github.com/fakenewsresearch/dataset} - This link is currently private and will be made public after the paper's acceptance. It will be provided as supplementary material.}. This initiative offers researchers a centralized, comprehensive portal for accessing and analyzing relevant datasets, with regular updates. Due to page constraints, only a portion of the GitHub pages are displayed here. Table~\ref{tab:datasets} provides an overview of key datasets included in the repository.

\subsection{Types of data collected}
Fake news datasets are typically categorized into \textit{\textbf{textual}}, \textit{\textbf{visual}}, and \textit{\textbf{multimodal}} data types to better address the different forms of fake news content. For instance, textual datasets focus on fake news articles and posts, visual datasets include manipulated images and videos, and multimodal datasets combine both text and visual elements to provide a comprehensive analysis \cite{nakamura-etal-2020-fakeddit,zhang2024scenefnd}.

\paragraph{Textual datasets}
These datasets focus on written content, including articles, headlines, and social media posts. Examples include the LIAR~\cite{wang2017liar} and MisInfoText~\cite{torabi2019big} datasets, which are used for analyzing linguistic patterns and textual inconsistencies. Challenges include the need for context in short texts and detecting satire or sarcasm \cite{ali2023deep}.

\paragraph{Visual datasets}
Visual datasets consist of images and videos used to detect fake news through visual content analysis. The Verification Corpus~\cite{boididou2018detection} and FCV-2018~\cite{papadopoulou2019corpus} datasets help develop algorithms for identifying image manipulations and verifying video authenticity.

\paragraph{Multimodal datasets}
These datasets combine text, images, and videos for a comprehensive approach to fake news detection. Examples include FakeNewsNet~\cite{shu2018fakenewsnet} and r/fakeddit~\cite{nakamura-etal-2020-fakeddit}, which improve detection accuracy by cross-referencing multiple data types.

\paragraph{Generative machine text datasets}
A recent trend in fake news detection focuses on generative machine text datasets, including those produced by AI models like GPT-3. These datasets are crucial for understanding and detecting AI-generated misinformation. State-of-the-art methods for detecting fake news in machine-generated text achieve over 90\% accuracy in controlled environments, but their applicability in real-world settings remains challenging \cite{valiaiev2024detection}. The M4 dataset~\cite{wang-etal-2024-m4} exemplifies this trend, containing both human-written and machine-generated text, aiding the development of distinguishing tools.

\subsection{Common Features and Labels}
Fake news datasets typically include a variety of features such as linguistic, metadata, and visual features, with labels categorizing content as fake or real. In datasets that focus exclusively on images, visual features like image metadata, pixel patterns, and manipulation traces are used to distinguish between authentic and fake images. These visual datasets are essential for identifying visually deceptive information, which cannot be assessed through textual analysis alone \cite{kondamudi2023comprehensive}.

\paragraph{Linguistic Features}
These features include syntax, semantics, and stylistics derived from text. They are crucial for training machine learning models to classify textual content~\cite{shu2018fakenewsnet, shu2019beyond, garg2022linguistic, choudhary2021linguistic}. Examples include n-grams, part-of-speech tags, and sentiment scores~\cite{chakraborty2016stop, garg2022linguistic, choudhary2021linguistic}.

\paragraph{Metadata}
Metadata provides contextual information such as publication date, author, source, and social media engagement metrics. It helps in assessing source credibility and tracking information spread. The PHEME dataset~\cite{zubiaga2016analysing} utilizes metadata to analyze rumor spread on Twitter.

\paragraph{Visual Features}
These features include image metadata, pixel data, and patterns identified through image processing. Advances in computer vision, such as convolutional neural networks, enhance the detection of visual inconsistencies~\cite{jin2017multimodal}.

\paragraph{Labels}
Labels generally classify content as fake or real, with some datasets offering more detailed labels indicating degrees of falsity. Binary labels are simple, while nuanced labels offer deeper insights into disinformation \cite{wang2017liar}. Table~\ref{tab:rating_scales} illustrates the various rating scales used in different datasets. For example, the CREDBANK dataset uses a five-point scale from "Certainly Inaccurate" to "Certainly Accurate," while the PHEME dataset categorizes information as "true," "false," or "unverified."

\begin{table*}[htbp]
\centering

\begin{tabular}{ll}
\hline
\textbf{Dataset} & \textbf{Rating Scale} \\
\hline
CREDBANK~\cite{mitra2015credbank} & 5 values (Cert., Prob. Inacc., Doubt., Prob. Acc., Cert.) \\
PHEME~\cite{zubiaga2016analysing} & 3 values (true, false, unverif.) \\
FacebookHoax~\cite{tacchini2017some} & 2 values (hoaxes, non-hoaxes) \\
LIAR~\cite{wang2017liar} & 6 values (pants-fire, false, barely-true, half-true, mostly-true, true) \\
BuzzFeed~\cite{horne2017just}  & 4 values (mostly true, not factual, mix, mostly false) \\
BuzzFace~\cite{santia2018buzzface} & 4 values (mostly true, mostly false, mix, no factual) \\
FakeNewsNet~\cite{shu2018fakenewsnet} & 2 values (fake, real) \\
Yelp~\cite{barbado2019framework} & 2 values (fake, trustful) \\
MisInfoText~\cite{torabi2019big} & 4 (BuzzFeed), 5 (Snopes) values \\
NELA-GT-2018~\cite{norregaard2019nela} & 2 values (true, false) \\
FCV-2018~\cite{papadopoulou2019corpus}  & 2 values (true, false) \\
Verification Corpus & 2 values (true, false) \\
r/fakeddit~\cite{nakamura-etal-2020-fakeddit} & 5 values (Cert., Prob. Inacc., Doubt., Prob. Acc., Cert.) \\
M4~\cite{wang-etal-2024-m4} & 5 values (Cert., Prob. Inacc., Doubt., Prob. Acc., Cert.) \\
\hline
\end{tabular}
\caption{\label{tab:rating_scales}
Summary of datasets and their rating scales.}
\end{table*}

\paragraph{Source}
Datasets are collected from social media platforms, news websites, and fact-checking organizations. Each source presents unique challenges, such as the dynamic nature of social media and the credibility of news websites \cite{shu2018fakenewsnet}.

\paragraph{Annotation Process}
Annotation methods include manual, crowdsourced, and automated approaches. Manual annotation, exemplified by the LIAR dataset \cite{wang2017liar}, guarantees high accuracy but is labor-intensive and time-consuming. On the other hand, crowdsourcing, as utilized by BuzzFeed \cite{horne2017just}, and automation, as implemented in FakeNewsNet \cite{shu2018fakenewsnet}, provide scalable solutions, though they may occasionally compromise on quality \cite{ghanem2018stance}.

\paragraph{Language}
Multilingual datasets, including M4 \cite{wang-etal-2024-m4}, are vital for ensuring global applicability, though they demand language-specific models and complex data processing \cite{baly2018predicting}. Furthermore, the \href{https://github.com/Arko98/TALLIP-FakeNews-Dataset}{TALLIP-FakeNews-Dataset} serves as a multilingual resource that encompasses low-resource languages \cite{10.1145/3472619talipp}.

Understanding these characteristics helps researchers select appropriate datasets, ensuring their methodologies align with the dataset's features and constraints. Table \ref{tab:rating_scales} outlines the rating scales employed across different datasets, while Tables \ref{tab:dataset_sizes} and \ref{tab:datasets} provide detailed information on the year of publication, language coverage, data types, and access methods. This comprehensive overview highlights the diversity and scope of available datasets, assisting researchers in selecting the most appropriate datasets for their specific needs. The trend towards multimodal and generative machine text datasets reflects the evolving landscape of fake news detection and underscores the necessity for advanced analytical methods.

\begin{table*}[htbp]
\centering

\begin{tabular}{llccc}
\hline
\textbf{Dataset} & \textbf{Year} & \textbf{Language} & \textbf{Type} & \textbf{Availability}\\
\hline
CREDBANK~\cite{mitra2015credbank} & 2015 &{\color{blue} \ding{108}} & {\color{green} \ding{110}} & \href{https://github.com/compsocial/CREDBANK-data}{\faUnlock} \\
PHEME~\cite{zubiaga2016analysing} & 2016 & {\color{blue} \ding{108}} {\color{yellow} \ding{108}} & {\color{green} \ding{110}} {\color{yellow} \ding{110}} & \href{https://figshare.com/articles/PHEME_rumour_scheme_dataset_journalism_use_case/2068650/2}{\faUnlock}\\
FacebookHoax~\cite{tacchini2017some} & 2017 & {\color{blue} \ding{108}} & {\color{green} \ding{110}} & \href{https://github.com/gabll/some-like-it-hoax}{\faUnlock} \\
LIAR~\cite{wang2017liar} & 2017 & {\color{blue} \ding{108}} & {\color{green} \ding{110}} & \href{https://paperswithcode.com/dataset/liar}{\faUnlock} \\
BuzzFeed~\cite{horne2017just} & 2017 & {\color{blue} \ding{108}} & {\color{green} \ding{110}} & \href{https://github.com/BuzzFeedNews/2016-10-facebook-fact-check/tree/master/data}{\faUnlock} \\
BuzzFace~\cite{santia2018buzzface} & 2018 & {\color{blue} \ding{108}} & {\color{green} \ding{110}} & \href{https://github.com/gsantia/BuzzFace}{\faUnlock} \\
FakeNewsNet~\cite{shu2018fakenewsnet} & 2018 & {\color{blue} \ding{108}} & {\color{green} \ding{110}} {\color{red} \ding{110}} & \href{https://github.com/KaiDMML/FakeNewsNet}{\faUnlock} \\
Yelp~\cite{barbado2019framework} & 2019 & {\color{blue} \ding{108}} & {\color{green} \ding{110}} & \href{mailto:o.araque@upm.es}{\Letter} \\
MisInfoText~\cite{torabi2019big} & 2019 & {\color{blue} \ding{108}} & {\color{green} \ding{110}} & \href{https://github.com/sfu-discourse-lab/MisInfoText}{\faUnlock} \\
NELA-GT-2018~\cite{norregaard2019nela} & 2019 & {\color{blue} \ding{108}} & {\color{green} \ding{110}} & \href{https://dataverse.harvard.edu/dataset.xhtml?persistentId=doi:10.7910/DVN/YHWTFC}{\faUnlock} \\
FCV-2018~\cite{papadopoulou2019corpus} & 2019 & {\color{blue} \ding{108}} {\color{red} \ding{108}} {\color{green} \ding{108}} {\color{orange} \ding{108}} {\color{yellow} \ding{108}} {\color{purple} \ding{108}} {\color{pink} \ding{108}} {\color{brown} \ding{108}} & {\color{green} \ding{110}} {\color{black} \ding{110}} & \href{https://mklab.iti.gr/results/fake-video-corpus/}{\faUnlock} \\
Verification Corpus~\cite{boididou2018detection} & 2019 & {\color{blue} \ding{108}} {\color{green} \ding{108}} {\color{red} \ding{108}} {\color{purple} \ding{108}} & {\color{green} \ding{110}} {\color{red} \ding{110}} & \href{https://github.com/MKLab-ITI/image-verification-corpus}{\faUnlock} \\
r/fakeddit~\cite{nakamura-etal-2020-fakeddit} & 2020 & {\color{blue} \ding{108}} & {\color{green} \ding{110}} {\color{red} \ding{110}} {\color{yellow} \ding{110}} & \href{https://fakeddit.netlify.app/}{\faUnlock} \\
M4~\cite{wang-etal-2024-m4} & 2024 & {\color{blue} \ding{108}} {\color{cyan} \ding{108}} {\color{black} \ding{108}} {\color{magenta} \ding{108}} {\color{gray} \ding{108}} {\color{orange} \ding{108}} & {\color{green} \ding{110}} {\color{magenta} \ding{110}} & \href{https://github.com/mbzuai-nlp/M4?tab=readme-ov-file#data}{\faUnlock} \\
\hline
\end{tabular}
\caption{  \label{tab:datasets}
Summary of Various Datasets with Language and Availability Information. Languages $\rightarrow$ English: {\color{blue} \ding{108}}, Russian: {\color{red} \ding{108}}, Spanish: {\color{green} \ding{108}}, Arabic: {\color{orange} \ding{108}}, German: {\color{yellow} \ding{108}}, Catalan: {\color{purple} \ding{108}}, Japanese: {\color{pink} \ding{108}}, Portuguese: {\color{brown} \ding{108}}, Dutch: {\color{black} \ding{108}}, French: {\color{purple} \ding{108}}, Chinese: {\color{cyan} \ding{108}}, Urdu: {\color{magenta} \ding{108}}, Indonesian: {\color{gray} \ding{108}}, Arabic: {\color{orange} \ding{108}}. Type $\rightarrow$ Text: {\color{green} \ding{110}}, Images: {\color{red} \ding{110}}, Video: {\color{black} \ding{110}}, Generated-text: {\color{magenta} \ding{110}}, Metadata: {\color{yellow} \ding{110}}.
}
\end{table*}

\section{Impact of dataset properties on detection algorithms}

\subsection{Performance influence}
The performance of detection algorithms is significantly influenced by dataset characteristics. Larger datasets generally improve classification performance by providing more information, aiding in pattern generalization during training. In contrast, smaller datasets often lead to overfitting and less reliable models due to limited variability and detail \cite{gravanis2019behind, ahmad2019comparison, shu2019beyond, faustini2020fake, dhawan2022fakenewsindia}. For instance, the NELA-GT-2018 dataset, with approximately 713,000 articles, enhances model performance by offering diverse data, reducing overfitting, and improving generalization. Conversely, smaller datasets like the Verification Corpus, with about 15,630 articles, present challenges such as overfitting despite being relatively substantial.

\subsection{Specific properties leading to better performance}
Certain dataset characteristics consistently improve detection accuracy and robustness \cite{shu2019role, wang2020weak}.

\paragraph{High-quality annotations} Detailed, high-quality annotations provide richer information, resulting in more accurate predictions. Datasets that accurately represent the original distribution tend to yield better performance regardless of size. Incorporating diverse features, such as numerical and textual data, enhances generalization. For example, the LIAR dataset \cite{wang2017liar} includes extensive fact-checking data with multiple truthfulness levels, allowing models to learn subtle distinctions and improve accuracy in classifying statement veracity. For additional information on rating scales, see Table \ref{tab:rating_scales}.

\paragraph{Balanced class distribution} Datasets with balanced class distributions help train unbiased models, reducing the risk of favoring a particular class. Balanced distributions enhance performance across various models. For example, the PHEME dataset \cite{zubiaga2016analysing} includes a balanced distribution of rumor and non-rumor data, ensuring the model does not become biased toward one class, leading to more robust and reliable rumor detection.

\section{Challenges and limitations in current fake news datasets}
\subsection{Common challenge}
Fake news detection models face several significant challenges related to the datasets used:
\paragraph{Data imbalance} A major issue is the disproportionate number of real news instances compared to fake news instances within datasets. This imbalance can bias models towards predicting real news, undermining their ability to effectively identify fake news \cite{murayama2021dataset, lai2024rumorllm}.
\paragraph{Noise in data} Datasets often contain noisy data, which includes irrelevant or misleading information. This noise can come from various sources, such as user-generated content that may include typos, slang, or inconsistent information, making it difficult for models to learn accurate patterns \cite{d2021fake, murayama2021dataset}.
\paragraph{Dynamic nature of fake news} Fake news evolves constantly, with new topics, narratives, and formats emerging regularly. Due to distribution drift, this rapid evolution requires continual updates to datasets to ensure they accurately reflect the current landscape of misinformation. Models trained on outdated datasets may not perform well on new types of fake news \cite{murayama2021dataset, khan2023fakewatch}.

\subsection{Biases in datasets}
Biases present in fake news datasets can significantly impact detection outcomes:

\paragraph{Selection bias} This bias occurs when the data collected does not accurately represent the broader population of fake news. For instance, datasets might over-represent certain types of fake news while under-representing others, leading to models that perform well on some types but poorly on others \cite{murayama2021dataset,d2021fake}.
\paragraph{Labelling bias} The process of labeling news as fake or real can introduce bias, especially if the annotators have preconceptions or if the labeling criteria are inconsistent. This issue is particularly pronounced in crowdsourced annotations, which can lead to varied interpretations and inconsistencies \cite{murayama2021dataset,d2021fake}.
\paragraph{Cultural and linguistic bias} Datasets predominantly in one language or from a specific cultural context may not generalize well to other languages or cultures. This limits the applicability of models trained on such datasets to detect fake news in a global context \cite{murayama2021dataset, khan2023fakewatch, d2021fake}.

\subsection{Influence on model performance}
These challenges and biases directly affect the performance of fake news detection models:

\paragraph{Reduced accuracy} Issues like data imbalance and noise can lead to lower accuracy in detection models. These models might become overfitted to the majority class (real news) and fail to accurately identify fake news, resulting in higher false negative rates \cite{murayama2021dataset,d2021fake}.

\paragraph{Inconsistent performance} Biases in the datasets can cause models to perform inconsistently across different types of fake news. A model trained on a biased dataset may excel in detecting certain narratives but fail in others, depending on the diversity and representativeness of the training data \cite{murayama2021dataset,d2021fake}.

\paragraph{Overfitting and underfitting} Noise and the evolving nature of fake news content can lead to overfitting to outdated or irrelevant patterns, or underfitting due to failure to capture the current state of misinformation. This affects the robustness and adaptability of the models \cite{lai2024rumorllm, murayama2021dataset}.

\subsection{Generalizability issues}

Dataset characteristics also influence how well models generalize across different datasets:

\paragraph{Domain specificity} Models trained on domain-specific datasets (e.g., social media posts) may not perform well on datasets from different domains (e.g., news articles). The features and patterns in one domain can be significantly different from those in another \cite{abdali2022multi, chen2022cross}.

\paragraph{Cross-lingual generalizability} Models trained on datasets in a particular language may struggle to generalize to other languages due to linguistic differences, limiting their effectiveness in multilingual or international contexts \cite{khan2023fakewatch, murayama2021dataset}.

\paragraph{Temporal relevance} Due to the dynamic nature of fake news, datasets quickly become outdated. Models trained on older data may not generalize well to new fake news stories, reducing their effectiveness over time \cite{lai2024rumorllm, murayama2021dataset, choras2021advanced}.

\section{Role of multimodal datasets in fake news detection}

\subsection{Comparison with unimodal datasets}

Multimodal datasets, which combine text, images, and videos, generally outperform unimodal datasets in fake news detection. Research shows that models using multimodal data better capture context and detect inconsistencies, leading to improved accuracy. For instance, the Fakeddit dataset, integrating text and images, achieved 87\% accuracy with a CNN architecture, surpassing text-only methods \cite{nakamura-etal-2020-fakeddit, wang2023cross, segura2022multimodal}. Studies indicate that multimodal news classification can improve accuracy by up to 8.11\% compared to text-only classification \cite{wang2021n24news}. These results are supported by further research, underscoring the superiority of multimodal approaches over unimodal ones in detecting fake news \cite{zhou2024clip, sormeily2024mefand}.

\subsection{Challenges of multimodal datasets}
Creating and utilizing multimodal datasets involve several challenges:

\paragraph{Data collection and integration} Collecting data from various sources (e.g., social media, news articles, images) and integrating them into a unified dataset is complex and labor-intensive. Ensuring that the data from different modalities are synchronized and accurately linked is critical for effective analysis \cite{segura2022multimodal, wang2023cross, hangloo2022combating}.

\paragraph{Annotation complexity} Annotating multimodal datasets requires expertise in both textual and visual analysis. The process is more time-consuming and expensive compared to annotating unimodal datasets, as it involves reviewing and labeling multiple types of data \cite{segura2022multimodal, wang2023cross}.

\paragraph{Computational resources} Processing multimodal data demands significant computational resources. Models must handle large volumes of data and perform complex feature extraction and integration, which can be computationally intensive and require advanced hardware \cite{bayoudh2022survey, wang2023cross}.

\subsection{Advantages and disadvantages of multimodal datasets}
\paragraph{Advantages}
\begin{enumerate}
    \item \textbf{Improved detection accuracy} By leveraging multiple data types, multimodal approaches can capture a richer set of features and contextual information, leading to higher detection accuracy. This allows for a more comprehensive understanding of the news content and better identification of fake news \cite{nakamura-etal-2020-fakeddit, wang2023cross}.
    \item \textbf{Enhanced robustness} Multimodal models are generally more robust to different types of fake news. They can cross-verify information from different modalities, which reduces the likelihood of false positives and negatives \cite{segura2022multimodal, wang2023cross}.
    \item \textbf{Contextual understanding} Combining text and visual data allows models to understand the context better. For instance, a sensational headline paired with an equally sensational image can be more easily identified as fake news \cite{nakamura-etal-2020-fakeddit, murayama2021dataset}.
\end{enumerate}

\paragraph{Disadvantages}
\begin{enumerate}
    \item \textbf{Resource intensiveness} Multimodal approaches require more computational power and storage, which can be a barrier for smaller research teams or organizations with limited resources \cite{nakamura-etal-2020-fakeddit, wang2023cross, segura2022multimodal}.
    \item \textbf{Complexity in implementation} Developing and maintaining multimodal models is more complex compared to unimodal models. This complexity involves sophisticated data preprocessing, feature extraction, and model integration techniques \cite{segura2022multimodal, wang2023cross}.
    \item \textbf{Data availability} High-quality multimodal datasets are harder to come by. Collecting and curating large-scale datasets that include both textual and visual content is challenging and resource-intensive \cite{nakamura-etal-2020-fakeddit, murayama2021dataset}.
\end{enumerate}

\section{Best practices for creating high-quality fake news datasets}

\paragraph{Annotation methodologies} Reliable annotation methodologies are essential for high-quality fake news datasets. Expert annotators familiar with fake news nuances help maintain accuracy and consistency \cite{khan2023fakewatch,kim2021systematic}. Automated methods can assist but require human oversight to address subtleties missed by algorithms \cite{nagy2021improving, d2021fake}. Quality control measures, such as cross-validation by multiple annotators and periodic reviews, are critical for dataset integrity \cite{nagy2021improving, d2021fake}. For example, the NELA-GT-2018 dataset employs cross-verification to reduce bias and error \cite{khan2023fakewatch}.

\paragraph{Incorporating real-world dynamics} Continuously updating datasets with new data reflecting current trends is crucial to capturing evolving fake news patterns \cite{nagy2021improving, d2021fake}. This involves monitoring various information sources, including social media and news websites, to capture emerging narratives and misinformation tactics \cite{huang2023harnessing,kim2021systematic}. Integrating these real-world dynamics keeps datasets relevant and useful for training robust fake news detection models \cite{huang2023harnessing, nagy2021improving}. Using automated tools for data scraping and periodic manual updates can help maintain dataset relevance \cite{huang2023harnessing, nagy2021improving}.

\paragraph{Ensuring reliability and validity} Maintaining the reliability and validity of fake news datasets involves several strategies. First, constructing datasets from diverse sources minimizes bias and increases representativeness \cite{khan2023fakewatch,d2021fake}. Second, rigorous pre-processing steps, such as removing duplicates and normalizing data, enhance dataset quality \cite{khan2023fakewatch,d2021fake}. Third, standardized metrics and evaluation methods aid in assessing dataset performance and reliability \cite{khan2023fakewatch,d2021fake}. The FakeWatch ElectionShield dataset, for instance, consolidates data from multiple sources and employs robust pre-processing and annotation techniques to ensure high quality and validity \cite{khan2023fakewatch,d2021fake}.

\section{Evolution of fake news detection models with dataset availability}

\paragraph{Trends in model performance}
The performance and architecture of fake news detection models have significantly evolved with new datasets. Early models relied on textual data and basic machine learning algorithms. With the advent of sophisticated datasets, models now incorporate advanced techniques like deep learning, NLP, and multimodal approaches. Large-scale datasets like LIAR and FakeNewsNet have enabled complex models to analyze linguistic features, sentiment, and metadata \cite{wang2017liar, mishra2023comprehensive}. Transformer-based models like BERT and GPT have recently improved fake news detection accuracy and robustness.

\paragraph{Impact of dataset versions and updates}
Updated datasets profoundly impact longitudinal studies and model effectiveness. Regular updates ensure models remain relevant and handle the latest fake news trends and tactics. For example, NELA-GT-2020 and NELA-GT-2022 include additional data and updated annotations reflecting current misinformation trends, enhancing model effectiveness over time \cite{wang2017liar, mishra2023comprehensive}. Studies show that models trained on continuously updated datasets outperform those trained on static datasets, adapting better to the evolving nature of fake news. Periodic updates and new dataset versions mitigate biases and improve model generalizability across contexts and time periods.

\section{Ethical considerations in fake news datasets}

\paragraph{Privacy and consent}
Ensuring privacy and consent for individuals in fake news datasets is crucial. Anonymizing personal information prevents identification and misuse, using techniques like tokenization and removing PII. For instance, the FakeWatch ElectionShield dataset anonymizes user data to safeguard privacy \cite{khan2023fakewatch, kim2021systematic, huang2023harnessing}. Obtaining informed consent involves clearly communicating data usage and ensuring user agreement \cite{shushkevich2023improving}.

\paragraph{Societal impacts}
The societal impacts of using specific datasets for fake news detection must be carefully considered. Propagating biased or unverified information can influence public opinion and cause harm. Datasets focusing on particular fake news types or sources may inadvertently reinforce biases if not managed properly \cite{d2021fake, shushkevich2023improving}. Developing and deploying fake news detection models must be transparent and accountable to avoid perpetuating or creating biases. Ethical implications, such as influencing elections or public health decisions, highlight the need for rigorous standards in dataset creation and use \cite{wang2017liar, shushkevich2023improving}.

\section{Future directions}

\paragraph{Adaptive and evolving datasets} As fake news evolves, creating adaptive datasets to reflect changing patterns is essential. One approach is developing dynamic datasets regularly updated with new data from social media, news websites, and forums. Additionally, incorporating feedback mechanisms based on fake news detection model performance can maintain relevance and effectiveness. Using machine learning to identify and integrate emerging patterns further enhances adaptiveness.

\paragraph{Role of synthetic data generation} Synthetic data generation offers a way to supplement real-world datasets, addressing data scarcity and imbalance. Techniques like Generative Adversarial Networks (GANs) can create realistic fake news samples to augment datasets, training more robust models with diverse examples. However, challenges include ensuring authenticity, avoiding biases, and balancing synthetic with real data to maintain dataset integrity and reliability.
\section{Conclusion}

This survey has examined into the pivotal role of datasets in the development of effective fake news detection models. We explored various facets, including the characteristics and challenges of existing fake news datasets, their impact on model performance, and the importance of multimodal and high-quality datasets. Additionally, the survey highlighted significant ethical considerations in the creation and utilization of these datasets, along with the evolution of detection models in response to the availability of new and updated datasets.

Datasets are the cornerstone of advancements in fake news detection technologies. As the landscape of fake news continues to shift and evolve, creating adaptive and high-quality datasets becomes increasingly critical. Future research should focus on developing dynamic datasets that adapt to changing misinformation patterns and leveraging synthetic data to enhance diversity and robustness. Ensuring ethical standards in creating and using these datasets will be vital in maintaining public trust and the effectiveness of fake news detection efforts. The ongoing innovation and collaboration in this field are essential to addressing the complex and evolving challenges posed by fake news, ensuring that detection technologies can keep pace with new tactics and trends in misinformation.

\section*{Acknowledgment}

The authors would like to thank the European Union for funding this work under the Horizon Europe grant OMINO (grant number 101086321). The views and opinions expressed herein are solely those of the author(s) and do not necessarily reflect those of the European Union or the European Research Executive Agency. Neither the European Union nor the European Research Executive Agency bears responsibility for them.

\bibliographystyle{unsrt}  
\bibliography{references}

\end{document}